\documentclass{article}  %
\usepackage{iclr2026_conference,times}

\usepackage{amsmath,amsfonts,bm}

\def\eqref#1{equation~\ref{#1}}

\def\1{\bm{1}}

\DeclareMathAlphabet{\mathsfit}{\encodingdefault}{\sfdefault}{m}{sl}
\SetMathAlphabet{\mathsfit}{bold}{\encodingdefault}{\sfdefault}{bx}{n}

\usepackage{xspace}
\usepackage{xcolor}
\usepackage{enumitem}
\usepackage{amsmath,amssymb}
\usepackage{algorithm,algpseudocode,listings}
\usepackage{booktabs,multirow,adjustbox,diagbox,threeparttable,tabularray,setspace}
\usepackage{wrapfig}
\usepackage{graphicx}
\usepackage{subcaption}
\usepackage{tabularx}
\usepackage{array}
\usepackage{makecell}
\usepackage{longtable} 
\usepackage{paracol}
\usepackage{float}
\usepackage[tableposition=top]{caption}
\usepackage{colortbl}
\usepackage{color}
\usepackage{multicol}
\usepackage{lipsum}
\usepackage{graphicx}
\usepackage{nicematrix}
\usepackage{enumitem}
\usepackage{tabularx}
\usepackage{tabularx, array}
\newcolumntype{Y}{>{\centering\arraybackslash}X} 
\definecolor{citeblue}{rgb}{0.21,0.49,0.74}
\usepackage[pagebackref=false,breaklinks,colorlinks,citecolor=citeblue,bookmarks=false]{hyperref}
\usepackage{wrapfig}
\usepackage[capitalize]{cleveref}  %

\crefname{section}{Sec.}{Secs.}
\Crefname{section}{Section}{Sections}
\crefname{appendix}{Appendix}{Appendices}
\Crefname{appendix}{Appendix}{Appendices}
\crefname{table}{Table}{Tables}
\Crefname{table}{Table}{Tables}
\crefname{figure}{Fig.}{Figs.}
\Crefname{figure}{Figure}{Figures}
\crefname{equation}{Eq.}{Eqs.}
\Crefname{equation}{Equation}{Equations}
\crefname{theorem}{Thm.}{Thms.}
\Crefname{theorem}{Theorem}{Theorems}
\crefname{lemma}{Lem.}{Lems.}
\Crefname{lemma}{Lemma}{Lemmas}
\crefname{remark}{Rem.}{Rems.}
\Crefname{remark}{Remark}{Remarks}
\crefname{corollary}{Cor.}{Cors.}
\Crefname{corollary}{Corollary}{Corollaries}
\crefname{algorithm}{Alg.}{Algs.}
\Crefname{algorithm}{Algorithm}{Algorithms}
\usepackage{titletoc}
\usepackage{hyperref}
\definecolor{citecolor}{HTML}{001dc6}
\definecolor{pink}{HTML}{ed008c}
\hypersetup{
    colorlinks,
    linkcolor=citecolor,
    urlcolor=pink,
    citecolor=citecolor
}

\definecolor{darkblue}{RGB}{0,0,139}
\definecolor{Ourmethod}{RGB}{77, 107, 254}
\definecolor{orange}{RGB}{255,151,0}
\definecolor{cellred}{RGB}{213, 123, 101}
\definecolor{cellgreen}{RGB}{0, 205, 0}
\definecolor{cellblue}{RGB}{54, 125, 189}
\definecolor{codegreen}{rgb}{0,0.6,0}
\definecolor{codegray}{rgb}{0.5,0.5,0.5}
\definecolor{codepurple}{rgb}{0.58,0,0.82}
\definecolor{backcolour}{rgb}{1.0,1.0,1.0}
\lstdefinestyle{mystyle}{
    backgroundcolor=\color{backcolour},
    commentstyle=\color{codegreen},
    keywordstyle=\color{magenta},
    numberstyle=\tiny\color{codegray},
    stringstyle=\color{codepurple},
    basicstyle=\ttfamily\scriptsize,
    breakatwhitespace=false,
    breaklines=true,
    captionpos=b,
    keepspaces=true,
    numbers=left,
    numbersep=5pt,
    showspaces=false,
    showstringspaces=false,
    showtabs=false,
    tabsize=2
}
\usepackage[most,skins,theorems]{tcolorbox}
\usepackage{multirow}

\usepackage{rotating}  %

\newcolumntype{C}[1]{>{\centering\arraybackslash}p{#1}}
\newcolumntype{L}[1]{>{\arraybackslash}p{#1}}

\definecolor{demphcolor}{gray}{.2}

\definecolor{demphcolorinline}{gray}{.3}

\definecolor{demphcolor1}{gray}{.6}

\newcommand{\tocite}[1]{{\color{red} [TO CITE]}}

\newcommand{\ours}{FinSight}

\title{FinSight: Towards Real-World Financial Deep Research}

\author{%
  \textbf{Jiajie Jin}$^{1*}$,
  \textbf{Yuyao Zhang}$^{1*}$,
  \textbf{Yimeng Xu}$^{1}$,
   \textbf{Hongjin Qian}$^{2}$,
   \textbf{Yutao Zhu}$^{1}$,
  \textbf{Zhicheng Dou}$^{1\dagger}$
  \\
  $^1$Gaoling School of Artificial Intelligence, Renmin University of China\\
  $^2$BAAI\\
  $^*$Equal Contribution, $^{\dagger}$Corresponding author \\
  \texttt{\{jinjiajie, dou\}@ruc.edu.cn} \\
}

\iclrfinalcopy

\begin{document}

\maketitle

\begin{abstract}
Generating professional financial reports is a labor-intensive and intellectually demanding process that current AI systems struggle to fully automate. To address this challenge, we introduce \ours{} (\textbf{Fin}ancial In\textbf{Sight}), a novel multi-agent framework for producing high-quality, multimodal financial reports.
The foundation of \ours{} is the Code Agent with Variable Memory (CAVM) architecture, which unifies external data, designed tools, and agents into a programmable variable space, enabling flexible data collection, analysis and report generation through executable code. To ensure professional-grade visualization, we propose an Iterative Vision-Enhanced Mechanism that progressively refines raw visual outputs into polished financial charts. Furthermore, a Two-Stage Writing Framework expands concise Chain-of-Analysis segments into coherent, citation-aware, and multimodal reports, ensuring both analytical depth and structural consistency.
Experiments on various company and industry-level tasks demonstrate that \ours{} significantly outperforms all baselines, including leading deep research systems in terms of factual accuracy, analytical depth, and presentation quality, demonstrating a clear path toward generating reports that approach human-expert quality.
\end{abstract}
\section{Introduction}

Investment decisions worth billions of dollars hinge on the quality and timeliness of financial research reports~\citep{template_financial_report}. These reports translate raw market data into strategic insights, serving as analytical support for asset managers, equity researchers, and institutional investors. However, producing such reports remains a challenging task due to the overwhelming volume of financial data and the demand for rapid, high-quality analysis~\citep{neural_report_generation, financial_report_chunking}.
Recent advances in artificial intelligence, particularly in reasoning models \citep{2409_openai_o1, deepseek-r1, kimi_k2}, deep search and research applications~\citep{openai_deep_research, gemini_deep_research,grok_deepsearch,opendeepresearch}, present great potential in solving these labor-intensive collecting and analyzing tasks. Despite these technical advances, significant challenges persist in automating the generation of full financial research reports that meet the high standards for data accuracy, analytical depth, and multimodal content integration~\citep{yang2025multimodaldeepresearchergeneratingtextchart}.

Existing methods face several limitations that hinder their practical adoption:
\textbf{(1) Lack of Financial Domain Knowledge:} 
Most current systems, whether closed-source ~\citep{openai_deep_research,grok_deepsearch,gemini_deep_research} or open-source~\citep{hu2025owl,Li2025WebThinker}, are designed for general search scenarios, ignoring the integration of real-time heterogeneous financial data (both unstructured articles, news, and structured data).
\textbf{(2) Limited Multimodal Support and Visualization:} Almost all current methods can only produce plain-text reports, lacking diverse visualizations (\textit{e.g.}, figures, charts and tables) that are critical in conveying information \citep{yang2025multimodaldeepresearchergeneratingtextchart}.
\textbf{(3) Insufficient Analytical Depth:} Current methods often rely on rigid, predefined workflows for single-pass data collection~\citep{ircot,searcho1,jiajie_hira} and report generation~\citep{knowledge_report_generation}, preventing them from dynamically adjusting research strategies based on intermediate findings, ultimately limiting the analytical depth and insight of the final report.

To address these challenges, we introduce \textbf{\ours{}}, a novel multi-agent system that simulates the cognitive processes and analytical workflows of expert financial researchers. \ours{} operates three necessary stages: (1) Data Collection, which gathers up-to-date heterogeneous data and organizes it into a structured multimodal memory. (2) Data Analysis, where an interactive environment enables multi-round interactions with data, tools, and agents to derive a concise Chain-of-Analysis sequence. (3) Report Generation, which follows a draft outline to transform the data and Chain-of-Analysis into a formatted financial report with chart and data references, finally rendered in a professional style.

To realize \ours{}, we reconstruct the deep research workflow and propose a novel agent architecture, \textbf{Code Agent with Variable Memory (CAVM)}, where all data, tools, and agents are unified into a programmable variable space accessible and manipulable through executable code. This architecture leverages the code capabilities of language models~\citep{wang2024codeact, code_generation_survey, codeagent}, and enables flexible, scalable task handling from bottom-up data operations to high-level workflow orchestration.

To address the critical challenges of multimodal generation and analytical depth, we introduce two specialized mechanisms. To overcome the shortcomings of automated visualization, we propose an Iterative Vision-Enhanced Mechanism, where a vision-language model provides critical feedback to iteratively refine code-generated charts until they meet professional standards. For the challenge of generating coherent, long-form reports, we employ a Two-Stage Writing Framework. This framework first distills insights into concise Chain-of-Analysis segments, which then serve as a structured foundation for the Report Generation Agent to compose a full, context-aware report with tightly integrated visualizations and citations. Our extensive evaluations demonstrate that this synergistic approach enables \ours{} to significantly outperform existing methods, delivering reports with superior accuracy, depth, and multimodal coherence.

To comprehensively evaluate our method, we construct a high-quality benchmark featuring research tasks at both the company and industry levels, spanning multiple markets and diverse industries. Experiments demonstrate that our method significantly surpasses various deep research systems across three key dimensions: \textbf{Factual Accuracy}, \textbf{Analytical Depth}, and \textbf{Presentation Quality}, validating that \ours{} can generate rich, insightful, and multimodal financial research reports that approach the quality of human experts.

Our core contributions are as follows:

\begin{enumerate}
\item A Multi-Agent System for financial analysis based on the \textbf{Code Agent with Variable Memory (CAVM) architecture}, which integrates data, tools, and different agents into a unified programmable variable space, enabling flexible and scalable data collection, analysis, and report generation.

\item An \textbf{Iterative Vision-Enhanced Mechanism} for professional chart generation that integrates the code-generation capabilities of large language models with the visual understanding of vision–language models to iteratively refine basic charts into professional-quality visualizations.

\item A \textbf{Two-stage Writing Framework with Generative Retrieval} that progresses from short and concise Chain-of-Analysis segments to long and comprehensive financial reports, seamlessly integrating textual analysis with visual elements to meet the need for real-world financial multimodal deep research.
\end{enumerate}

\section{Method}
\begin{figure*}[!t]
    \centering
    \vspace{-2em}
    \includegraphics[width=\linewidth]{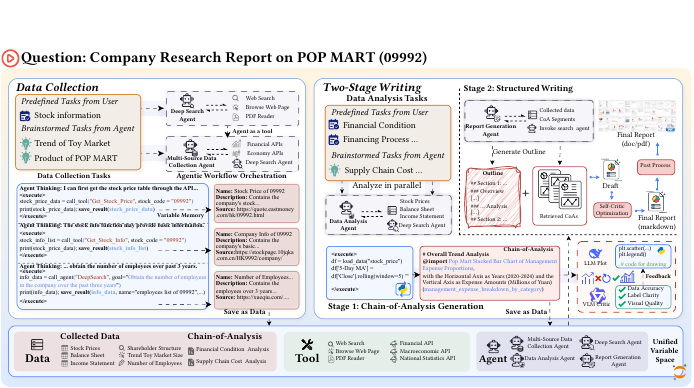}
    \caption{
    Overview of the \ours{} Framework.
    }
    \label{fig:intro_fig}
\end{figure*}

In this section, we give a definition of the multimodal financial report generation task (Section~\ref{section:problem_formulation}) and present the foundational Code Agent with Variable Memory (CAVM) architecture (Section~\ref{section:agent_description}), the Iterative Vision-Enhanced Mechanism (Section~\ref{section:vision_enhanced_mechanism}) and Two-Stage Writing Framework with Generative Retrieval (Section~\ref{section:writing}).

\subsection{Problem Formulation}
\label{section:problem_formulation}

We formalize the generation of open-domain multimodal financial research reports as follows. Given a research question $q$ (\textit{e.g.}, \textit{Research the development of the robotics industry}), the system is required to generate a structure report $R$:
\[
    R = \{r_1, r_2, \cdots, r_L\}, r_i \in \{T, V, C\},
\]
where:
\begin{itemize}[leftmargin=2em]
    \item \textbf{Texts:} $T = \{t_1, t_2, ..., t_n\}$ represents the textual analysis, such as executive summary, insights, conclusions and others.
    \item \textbf{Visualizations:} $V = \{v_1, v_2, ..., v_m\}$ denotes the visualizations, such as charts, graphs, and tables that can support textual analysis or add more information.
    \item \textbf{Citations:} $C = \{c_1, c_2, ..., c_k\}$ contains citations and references to data sources.
\end{itemize}

\subsection{The Framework of \ours{}}
\label{sec:finsight_frawework}

\ours{} is a multi-agent system designed to simulate the workflow of a professional financial analyst. The system realizes three core processes: multi-source data collection, multi-turn data analysis and progressive report writing, implemented through the CAVM architecture described in Section~\ref{section:agent_description}. The key design of this framework will be detailed in the following sections.

\paragraph{Data Collection}
To address the limitations of general web search systems in financial domains, we design two specialized agents for comprehensive data gathering: (1) \textbf{Deep Search Agent:} Conducts iterative, multi-round investigations using search engines and virtual browsers to gather comprehensive information with source verification. (2) \textbf{Multi-Source Data Collection Agent:} Collects heterogeneous data from financial databases, APIs, and web sources, leveraging different tools to access diverse information types. It can invoke the deep search agent for specific information requirements.
Instead of treating data collection as an isolated preliminary step, \ours{} allows the analysis and writing stages to dynamically invoke further data collection, ensuring broader and more relevant knowledge coverage.

\paragraph{Data Analysis}
Built on CAVM, the \textbf{Data Analysis Agent} executes analytical tasks via multi-turn code actions, dynamically deciding when to process data, invoke data collection workflows, or terminate with a concise Chain-of-Analysis (CoA) output (Section~\ref{section:writing}). It integrates the Iterative Vision-Enhanced Mechanism (Section~\ref{section:vision_enhanced_mechanism}) for professional chart generation.

\paragraph{Report Generation}
The \textbf{Report Generation Agent} handles drafting, optimization, and post-processing using the Two-Stage Writing Framework (Section~\ref{section:writing}). The process includes: (1) \textit{Drafting:} retrieving relevant CoA segments and structured data according to predefined outlines; (2) \textit{Self-reflective Optimization:} iteratively refining text for factual accuracy and consistency; and (3) \textit{Post-processing:} parsing identifiers, loading visualizations, formatting citations, and rendering into a publication-ready format.

\subsection{Code Agent with Variable Memory (CAVM)}
\label{section:agent_description}

\begin{wrapfigure}[18]{r}{0.60\textwidth}
	{
    \centering
    \vspace{-2ex}
	\includegraphics[width=80mm]{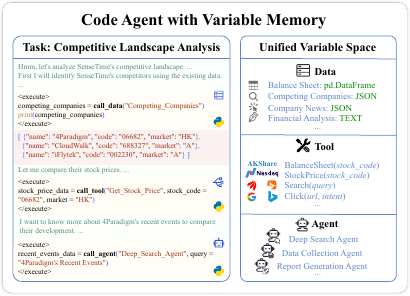}
        \caption{The design philosophy of CAVM architecture.}
 \label{fig:summary}
	}
\end{wrapfigure}

\paragraph{Motivation} In previous research, agents are often equipped with a role-specific toolkit, which limits flexible interaction between agents and the environment. Meanwhile, how to organize the intermediate data and memory of multi-agent system wisely is also underexplored. These factors significantly affect the overall performance of multi-agent systems when collaborating and handling complex tasks. Therefore, the core design philosophy of \textbf{Code Agent with Variable Memory (CAVM)} architecture is to empower each agent with autonomy, enabling flexible action space and seamless sharing of contextual memory across agents.

\paragraph{Unified Variable Space} The collaboration of multi-agent system involves diverse elements, which we abstract into three variable types, shown in Figure~\ref{fig:summary}: \textbf{data} ($\mathcal{V}_{data}$), including both structured and unstructured data variables; \textbf{tools} ($\mathcal{V}_{tool}$) with different functions; and \textbf{agents} ($\mathcal{V}_{agent}$) we designed with distinct role specification. In CAVM architecture, we represent all these variables in a unified variable space $\mathcal{V}$, 
\[
    \mathcal{V} = \mathcal{V}_{data} \cup \mathcal{V}_{tool} \cup \mathcal{V}_{agent}.
\] 
During multi-agent collaboration, the variable space is dynamically maintained and updated to support unified and efficient context management. 

\paragraph{Foundation Agent with Code Action} The agent operates in an iterative cycle. At each step, it first generates a reasoning trace and then produces executable code. The feedback from the code's execution then informs the planning for the subsequent step. This code-centric design leverages the language model's inherent coding abilities to unify diverse operations into a single, flexible, and scalable programming paradigm. Formally, at a given step $t$, the generation process can be decomposed as:
\begin{equation*}
\label{eq:agent_generation}
P_\theta(\mathcal{R}_t, \mathcal{C}_t \mid q, \mathcal{V}_{t-1}, \mathcal{H}_{t-1}) = \underbrace{P_\theta(\mathcal{R}_{t} \mid \Phi(\mathcal{V}_{t-1}), \cdot)}_{\text{Reasoning Process}} \cdot  \underbrace{P_\theta(C_{t} \mid  \mathcal{R}_t, \Phi(\mathcal{V}_{t-1}, C_{t-1}), \cdot)}_{\text{Code Action Generation}},
\end{equation*}
where $\mathcal{R}_t$, $\mathcal{C}_t$ and $\mathcal{H}_t$ respectively represent the generated reasoning chain, code snippet and interaction history of step $t$. $\Phi$ represents the format function that converts variable environment information into readable strings, with the following definition: 
\begin{equation}
\label{eq:format_details}
\Phi(\mathcal{V}_t) = \begin{cases}
\text{Info}(\mathcal{V}_0) & \text{if } t = 0, \\
\Phi(\mathcal{V}_{t-1}) \oplus \text{Info}(\mathcal{V}_t \setminus \mathcal{V}_{t-1}) & \text{if } t > 0.
\end{cases}
\end{equation}
Then, $\mathcal {C}_t $will be sent to the code interpreter for execution, and $V_ {t-1}$ will be modified and the result will be added to $H_ {t-1}$:
\begin{align}
\mathcal{V}_t, \text{output}_t &= \text{Execute}(\mathcal{C}_t, \mathcal{V}_{t-1}), \\
\mathcal{H}_t &= \mathcal{H}_{t-1} \oplus \text{output}_t.
\end{align}

\subsection{Iterative Vision-Enhanced Mechanism for Visualization}
\label{section:vision_enhanced_mechanism}

\textbf{Motivation}
Generating high-quality visualizations is a persistent challenge in automated report generation, particularly in data-intensive domains like finance that require nuanced analysis and presentation. Existing methods often rely on single-pass code execution or employ Vision-Language Models (VLMs) without incorporating visual feedback, which frequently leads to suboptimal outcomes. Drawing inspiration from Chain-of-Thought~\citep{COT} and Actor-Critic~\citep{PPO}, we propose a framework where an agent learns to progressively improve visualizations. This is achieved by iteratively plotting a chart and refining it based on critical feedback, ensuring both stable generation and continuous quality enhancement.

\textbf{Iterative Vision-Enhanced Mechanism } Specifically, the final output of the Data Analysis Agent includes the target chart specifications along with the corresponding descriptions and data. For each chart, the agent generates an initial visualization through executable plotting code, which is then evaluated by a VLM to give potential issues of visual cues (e.g. missing labels, inappropriate color schemes). These feedbacks are sent to the system, directing the iterative code generation until the output reaches professional quality.

\begin{equation*}
P(\mathcal{C}_{vis} \mid \mathcal{V}) = \prod_{t=1}^{M} P_\theta(\mathcal{C}_t^{vis} \mid \mathcal{C}_{t-1}^{vis}, \mathcal{F}_{t-1}, \mathcal{V}), \quad
\mathcal{F}_{t-1} = \text{VLM}(\text{Execute}(\mathcal{C}_{t-1}^{vis})),
\end{equation*}
where $M$ is the maximum number of iterations. The iteration continues until convergence or a predefined quality threshold is satisfied.

\subsection{Two-Stage Writing with Generative Retrieval}
\label{section:writing}

\begin{wrapfigure}[17]{r}{0.55\textwidth}
	{
    \centering
    \vspace{-2ex}
	\includegraphics[width=67mm]{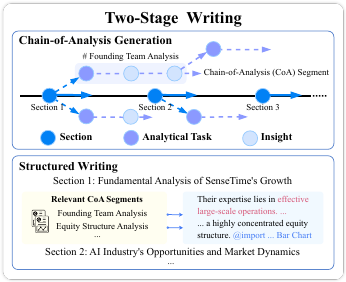}
        \caption{Chain-of-Analysis Illustration.}
 \label{fig:CoA}
	}
\end{wrapfigure}

\textbf{Motivation} A complete report encompasses analyses from multiple perspectives, which can be regarded as an integration of several Chains-of-Analysis. To generate long-form financial research reports with both textual depth and multimodal coherence, we design a \textbf{two-stage writing framework} augmented with generative retrieval. It decomposes the report writing process into (1) Chain-of-Analysis Generation and (2) Structured Writing with Generative Retrieval.

\textbf{Stage 1: Chain-of-Analysis Generation} 
Given the research question $q$, the Data Analysis Agent first generates a set of analytical perspectives $\mathcal{P} = \{p_1, p_2, ..., p_K\}$. The agent then performs parallel data analysis for each $p_i$, producing corresponding Chain-of-Analysis (CoA) that capture insights from distinct viewpoints.

Each CoA is generated based on the interaction history $\mathcal{H}_i$, accumulated during the data analysis process. To ensure coherence between textual content and referenced elements (e.g. figure, reference), this process is augmented with a generative retrieval mechanism that jointly produces textual contents along with element identifiers. These identifiers specify chart and reference attributes using natural language descriptions, enabling unified autoregressive generation. 
The process can be formalized as:
\begin{equation*}
    P(\mathcal{A} \mid q, \mathcal{V}) = P(\mathcal{P} \mid q, \mathcal{V}) \cdot \prod_{i=1}^{\mid \mathcal{P} \mid}P(a_i \mid p_i, \mathcal{V}).
\end{equation*}

\textbf{Stage 2: Structured Writing} Building on CoAs, a Report Generation Agent first constructs a report outline $\mathcal{O} = \{o_1, o_2, ..., o_n\}$, and then writes each section sequentially. For each section $s_i$, the agent dynamically retrieves the most relevant data and CoA segments from the unified variable memory $\mathcal{V}$, formalized as:
\begin{equation*}
    P(R \mid \mathcal{A}, \mathcal{V}, q) = P(\mathcal{O} \mid \mathcal{A},q) \cdot  \prod_{i=1}^{n} P(A_{\text{selected}}^{(i)}, \mathcal{V_\text{selected}^{(i)}} \mid \mathcal{A},\mathcal{V}, \cdot) \cdot P(s_i | s_{<i},  A_{\text{selected}}^{(i)} , \mathcal{V_\text{selected}^{(i)}}, \cdot).
\end{equation*}
To prevent hallucination of non-existent references and figures, agent is instructed to follow the identifiers established in $\mathcal{A}$.
To ensure reference accuracy, the agent strictly follows the identifiers established during the stage 1.

\section{Experiments}

\subsection{Dataset and Evaluation Metrics}

Financial research report generation remains an under-explored problem lacking appropriate evaluation benchmarks and metrics. To address this gap, we construct a high-quality benchmark specifically designed for financial research report generation, comprising a dataset of research targets with corresponding professional institutional reports and a comprehensive set of automated evaluation metrics. Details can be found in Appendix~\ref{app:eval}.

\paragraph{Dataset.} Our dataset encompasses research targets at both company and industry levels. For company-level analysis, we curated a diverse list of companies from authoritative financial platforms, covering different markets, industry sectors, and market capitalizations. For industry-level analysis, we selected high-attention industries from these platforms as research targets. For all targets, we collected in-depth analysis reports authored by professional brokerage institutions as golden reference reports to facilitate evaluation of data accuracy and analytical quality.

To ensure the quality of golden reference reports, we applied stringent filtering criteria, selecting only reports exceeding 20 pages in length and containing more than 20 charts and visualizations. Following established practices in report generation research~\citep{autosurvey, surveygo, Li2025WebThinker}, and considering the substantial time and computational costs associated with report generation and evaluation, we collected 20 samples: 10 company-level and 10 industry-level targets.

\paragraph{Evaluation Metrics.} We design 9 automated evaluation metrics across three critical dimensions, each ranging from 0 to 10 points. Detailed description of each metric can be found in Appendix~\ref{app:eval}.

\textbf{(1) Factual Accuracy:} Measures the reliability and correctness of generated content through Core Conclusion Consistency (alignment with reference conclusions), Textual Faithfulness (proper citation support), and Text-Image Coherence (consistency between textual and visual elements).

\textbf{(2) Information Effectiveness:} Evaluates the analytical value delivered to investors via Information Richness (distinct information points), Coverage (proportion of key reference information captured), and Analytical Insight (critical analysis and forward-looking recommendations).

\textbf{(3) Presentation Quality:} Assesses professional standards through Structural Logic (organizational coherence), Language Professionalism (adherence to financial terminology), and Chart Expressiveness (effective visualization utilization and aesthetic quality).

\subsection{Baselines}

We compare \ours{} against multiple categories of baselines:

\textbf{LLMs with Search Tools:} We evaluate leading large language models directly combined with search tools for report generation, including OpenAI GPT-5~\citep{openai_gpt5}, DeepSeek-R1~\citep{deepseek-r1}, and Claude-4.1-Sonnet~\citep{claude}.

\textbf{Deep Research Agents:} We compare against state-of-the-art commercial deep research products, including Gemini-2.5-Pro Deep Research~\citep{gemini_deep_research}, Grok Deep Search~\citep{grok_deepsearch}, OpenAI Deep Research~\citep{openai_deep_research}, and Perplexity Deep Research\footnote{\url{https://www.perplexity.ai/?model_id=deep_research}}. Details of baseline implementations can be found in Appendix~\ref{app:baseline}.

\subsection{Implementation Details}

Our backbone model uses DeepSeek-V3, and during the writing phase, we employ DeepSeek-R1 with reasoning capabilities. The maximum input length is set to 81,920, and the maximum output length is set to 16,384. For search, we use the Google Search API, setting the region to China and retrieving the top 10 search results. For evaluation, we use the multimodal model Gemini-2.5-Pro as our evaluation model. Details can be found in Appendix~\ref{app:implementation}.

\subsection{Main Results}
\begin{table}[!t]
  \centering
  \small                     
  \setlength{\tabcolsep}{3.5pt}
  \caption{Overall evaluation results on financial report generation benchmark.
           \textbf{Bold} denotes the highest score in each column,
           \underline{Underlined} denotes the second highest.}
  \label{tab:main_results}

  \begin{tabularx}{\textwidth}{l *{10}{Y}}
    \toprule
    \multirow{2}{*}{\textbf{Model}}
      & \multicolumn{3}{c}{\textbf{Factual}}
      & \multicolumn{3}{c}{\textbf{Analytical}}
& \multicolumn{3}{c}{\textbf{Presentation}} & \multirow{2}{*}{\textbf{Avg.}} \\
    \cmidrule(lr){2-4}\cmidrule(lr){5-7}\cmidrule(lr){8-10}
      & \textbf{Cons.} & \textbf{Faith.} & \textbf{T-I.}
      & \textbf{Rich.} & \textbf{Cover.} & \textbf{Ins.}
      & \textbf{Logic} & \textbf{Lang.} & \textbf{Vis.} & \\
    \midrule
    \multicolumn{11}{c}{\emph{LLM with Search Tools}} \\
    \midrule
    GPT-5 w/ Search & 5.60 & 6.45 & 3.45 & 5.95 & 4.30 & 5.60 & 6.80 & 5.30 & 2.95 & 5.16 \\
    Claude-4.1-Sonnet w/ Search & 4.75 & 5.15 & 2.70 & 5.45 & 4.60 & 4.70 & 6.15 & 5.50 & 2.20 & 4.58 \\
    DeepSeek-R1 w/ Search & 6.05 & 4.95 & 2.75 & 7.25 & 6.45 & 7.05 & 7.20 & 6.85 & 2.10 & 5.63 \\
    \midrule
    \multicolumn{11}{c}{\emph{Deep Research Agent}} \\
    \midrule
    Grok Deep Search & 4.05 & 5.70 & 3.80 & 5.10 & 3.65 & 4.55 & 5.75 & 4.80 & 4.10 & 4.61 \\
    Perplexity Deep Research & 4.10 & 5.60 & 4.25 & 4.00 & 2.70 & 3.60 & 5.30 & 3.90 & 3.95 & 4.16 \\
    OpenAI Deep Research & 5.60 & \underline{7.45} & \underline{4.90} & 6.35 & 6.40 & 5.90 & 6.90 & 6.85 & \underline{4.65} & 6.11 \\
    Gemini-2.5-Pro Deep Research & \textbf{7.10} & 6.80 & 4.65 & \underline{7.45} & \underline{7.75} & \underline{7.85} & \underline{7.65} & \underline{7.85} & 4.25 & \underline{6.82} \\
    \midrule
    \rowcolor{Ourmethod!25}
    \textbf{\ours{} (ours)} & \underline{6.85} & \textbf{7.50} & \textbf{7.85} & \textbf{8.70} & \textbf{8.30} & \textbf{8.45} & \textbf{8.05} & \textbf{8.10} & \textbf{9.00} & \textbf{8.09} \\
    \bottomrule
  \end{tabularx}
\end{table}

Table~\ref{tab:main_results} presents the performance of \ours{} against two categories of baselines on the financial research report generation task.
Overall, \ours{} achieves the highest overall score (8.09), significantly outperforming all baselines, including closed-source commercial agents like Gemini Deep Research (6.82) and OpenAI Deep Research (6.11).
This result validates the effectiveness of our proposed multi-agent framework for crafting in-depth financial research reports.
In terms of factuality, \ours{} obtains the best scores in both the faithfulness of text citations and text-image consistency, demonstrating the efficacy of the identifier mechanism designed within our Chain-of-Analysis process.

A noteworthy observation is that the consistency score of our model (6.85) is slightly lower than that of Gemini Deep Research (7.10).
Case studies reveal that our method prioritizes comprehensive data acquisition to deliver deeper insights.
This approach leads it to uncover more data-driven findings, rather than generating simplified conclusions from conventional search-based methods. 
\begin{wraptable}{r}{0.5\textwidth}
  \centering
  \caption{Ablation studies of our key design.}
  \setlength\tabcolsep{1.2pt} 
  \label{tab:ablation_study}
  \begin{tabular}{lccc}
    \toprule
    \textbf{Method} & \textbf{Fact.} & \textbf{Ana.} & \textbf{Pres.} \\
    \midrule
    \rowcolor{Ourmethod!25}
    \ours{} & 7.0 & 7.9 & 8.0\\
    \midrule
    w/o Iter.. & 6.9 & 7.2 & 7.5 \\
    w/o 2-Stage. & 6.4 & 5.9 & 6.3\\
    w/o Dyn. & 5.9 & 5.7 & 6.4 \\
    \bottomrule
  \end{tabular}
\end{wraptable}
The superiority of our method is further reflected in the analytical quality of the reports.
\ours{} scores the highest in information richness, coverage of key information from professional reports, and insightfulness.
Regarding presentation quality, our system demonstrates a comprehensive lead in logic, language, and visualization.
It particularly excels in visualization (9.00), far surpassing other methods and showcasing the advanced multimodal presentation capabilities of our system.

\section{Ablation Studies} 
We conduct ablation studies to evaluate the contribution of our key components, with results summarized in Table \ref{tab:ablation_study}. Key findings are as follows: (1) Removing iterative VLM feedback for chart generation causes a significant decline in both Presentation Quality (from 8.0 to 7.5) and Analytical Quality (from 7.9 to 7.2). This is primarily because the writing process relies on analyzing the generated images, lower-quality visuals impede the ability to perform insightful analysis based on the charts. (2) Merging analysis and writing into a single process leads to a significant drop in analytical quality (from 7.9 to 5.9) and factual accuracy (from 7.0 to 6.4), demonstrating the effectiveness of our proposed two-stage, analyze-then-write strategy. (3) Eliminating dynamic search during the analysis and writing phases results in a significant performance drop across all dimensions, including Factual Accuracy (from 7.0 to 5.9) and Analytical Depth (from 7.9 to 5.7). This highlights the necessity of acquiring additional knowledge during these stages to ensure comprehensive and factually correct reports.

\subsection{Analysis}
\paragraph{Statistical Analysis of Generation Process.} 

\begin{wraptable}{r}{0.4\textwidth}
  \centering
  \small
  \vspace{-3.5ex}
  \caption{Statistics of our generation process. We analyze metrics at both the CoA level and the final report level.}
  \label{tab:statistic_result}
  \setlength\tabcolsep{1.2pt} 
  \begin{tabular}{lc}
    \toprule
    \textbf{Metric} & \textbf{Avg. Value} \\
    \midrule
    \multicolumn{2}{l}{\textit{\textbf{Chain of Analysis (CoA)}}} \\
    \# Tokens  & 2,761 \\ 
    \# Images & 5.3 \\
    \midrule
    \multicolumn{2}{l}{\textit{\textbf{Final Report}}} \\
    \# Fin. API Calls & 18.3 \\
    \# Search Queries & 983.2 \\
    \# Browse Pages & 469.8 \\
    \midrule
    \# CoA Segments & 17.6 \\
    \# Tokens & 62,586 \\
    \# Images & 51.2 \\
    \bottomrule
  \end{tabular}
  \vspace{-5ex}
\end{wraptable}

Table~\ref{tab:statistic_result} summarizes report statistics. Key findings: (1) Each CoA is a self-contained multimodal block, averaging 2,761 tokens and 5.3 images. (2) A report synthesizes about 17.6 CoAs, yielding 62,586 tokens and 51.2 images. (3) Incorporating deep search introduces richer knowledge, with 983.2 searches and 469.8 browsed pages per report.

\begin{figure*}[!t]
    \centering
    \includegraphics[width=0.95\linewidth]{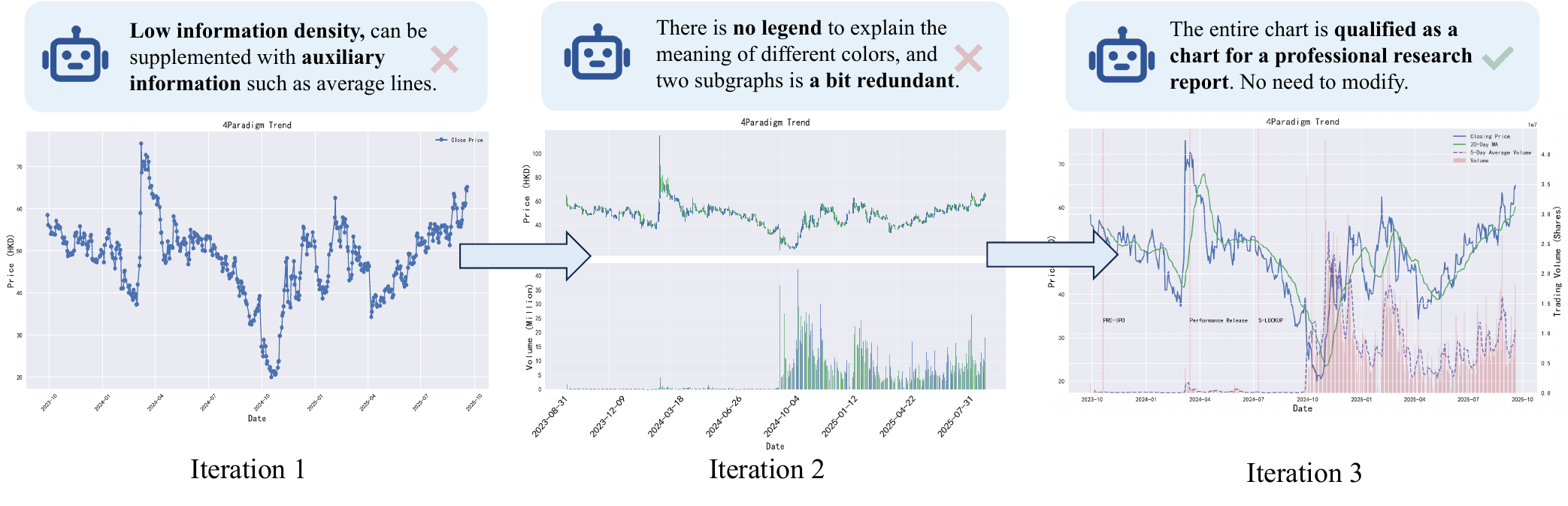}
    \caption{
    An example of our Iterative Vision-Enhanced Mechanism of Visualization. The chart is generated by matplotlib and seaborn package in Python.
    }
    \label{fig:iterative_vision}
\end{figure*}

\paragraph{Analysis of Image Generation.}As illustrated in Figure~\ref{fig:iterative_vision}, our Iterative Vision-Enhanced Mechanism progressively refines a stock chart over three iterations. In contrast to the initial, simplistic plot with low information density, the final visualization resolves this issue by integrating price and volume on a dual-axis, enriched with analytical overlays and contextual event markers, thereby presenting multifaceted data within a single view. This process is driven by critical VLM feedback across iterations, which targets improvements in aesthetics, information density, and other aspects.This suggests our mechanism is crucial for bridging the gap between automated chart generation and expert-quality financial visualizations.

\paragraph{Analysis of Report Length and Quality}
\begin{wrapfigure}[14]{r}{0.40\textwidth}
	{
    \vspace{-2ex}
    \centering
	\includegraphics[width=50mm]{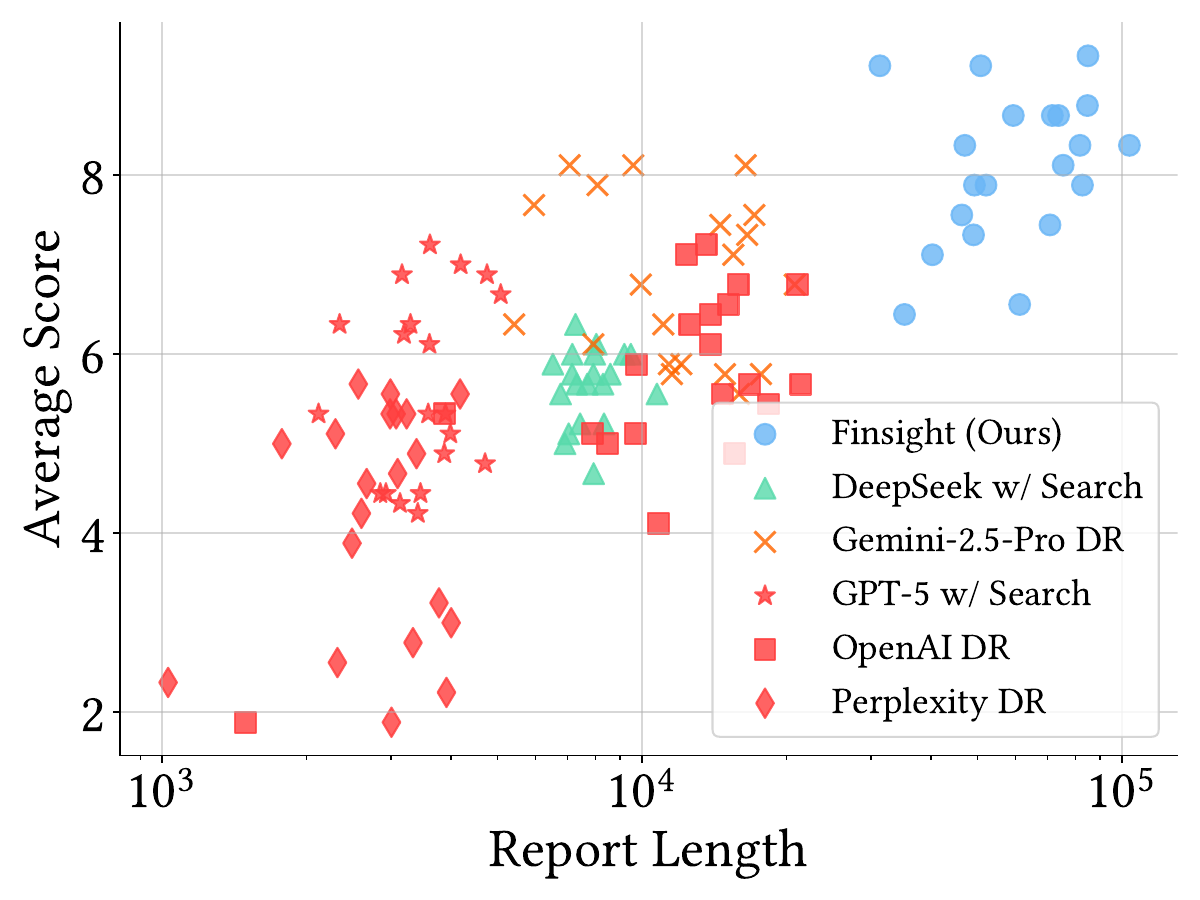}
    \caption{Correlation between report length and quality score across different methods.}
 \label{fig:length_score}
	}
\end{wrapfigure}
To further investigate the characteristics of the generated reports, we analyze the relationship between report length and overall quality score, as illustrated in Figure~\ref{fig:length_score}. The plot shows that the outputs from our method  are concentrated in the top-right quadrant, which indicates that our generated reports are not only comprehensive and of substantial length (typically over 20,000 words) but also of superior quality. We attribute this strong and consistent performance to our proposed two-stage writing framework. By first generating a concise Chain-of-Analysis, the model can then compose the final report based on richer, well-structured information, ensuring both analytical depth and coherence.

In contrast, baseline methods exhibit significant limitations. Simpler approaches like LLM with search tool, which often rely on single-pass generation, are typically constrained to shorter reports. Meanwhile, other deep research agents such as OpenAI DR and Perplexity DR display a wide scatter of data points across the plot, which signifies a critical lack of consistency. For these methods, a greater length does not reliably translate into higher quality, highlighting the effectiveness of our structured, two-stage approach.

\section{Related Work}

\subsection{Deep Research Systems}

Deep research systems represent a paradigm shift from traditional information retrieval to comprehensive knowledge synthesis, characterized by their ability to conduct multi-round information searching and integration. Current open-source deep research frameworks have emerged along several technical trajectories. 
ReAct-based agents \citep{react}, such as Open Deep Research \citep{openai_deep_research} and WebThinker \citep{Li2025WebThinker}, employ observation-thought-action loops with reasoning capabilities for iterative problem planning and execution. Multi-agent systems, including OWL \citep{hu2025owl} and Auto Deep Research \citep{autodeepresearch}, focus on collaborative problem-solving through agent specialization and coordination. Additionally, commercial systems represented by OpenAI Deep Research \citep{openai_deep_research} and Grok Deep Research \citep{grok_deepsearch} have demonstrated promising performance.
However, existing frameworks exhibit significant limitations in multimodal processing~\citep{yang2025multimodaldeepresearchergeneratingtextchart} and domain-specific applications~\citep{financial_report_chunking, template_financial_report}. Due to the text-centric design of report generation workflows and the base models' lack of native image generation capabilities~\citep{neural_report_generation,knowledge_report_generation}, current systems produce reports deficient in visual elements such as charts and diagrams. 
Furthermore, these systems demonstrate inadequate adaptation to financial domains, particularly in their inability to support for professional-grade chart generation, limited real-time market data integration, creating substantial gaps between system outputs and professional requirements.

\subsection{LLM Agents in Financial Domain}

Recent advances in Large Language Models have led to the development of various financial AI systems, each targeting specific aspects of financial analysis. Many of these works focus on stock price prediction and modeling \citep{zhang2025finworld, xiao2025tradingagentsmultiagentsllmfinancial} using multi-agent architectures. From a report generation perspective, FinTeam \citep{wu2025finteam} can provide analysis from multiple viewpoints including company and industry levels. However, due to its single-round generation process, the resulting analysis lacks depth and comprehensiveness. Similarly, FinRobot \citep{yang2024finrobot} directly inputs collected information to models for single-round investment recommendation generation.
Additionally, several open-source works~\citep{zhang2025finworld, template_financial_report} provide comprehensive tools and data interfaces, yet they lack well-designed frameworks for report generation. 
Overall, existing systems exhibit critical limitations for comprehensive financial research report generation, particularly regarding report depth, data breadth, and multimodal integration.

\section{Conclusion}
 In this paper, we introduce \ours{}, a multi-agent system designed to generate high-quality, multimodal financial research reports. At its core, \ours{} leverages the Code Agent with Variable Memory architecture, unifying data, tools, and agents into a single programmable space for dynamic, code-driven analysis. We also propose an Iterative Vision-Enhanced Mechanism to refine visualizations to professional standards and a Two-stage Writing Framework that expands concise analysis chains into comprehensive, coherent reports. Our experiments demonstrate that \ours{} significantly outperforms existing baselines across a range of financial research tasks. This work validates the effectiveness of our multi-agent, code-centric approach and highlights its potential to revolutionize the field of automated financial analysis.

\bibliographystyle{iclr2026_conference}
\bibliography{reference}
\newpage
\clearpage
\appendix

\section*{Appendix}
\startcontents[sections]
\printcontents[sections]{l}{1}{\setcounter{tocdepth}{3}}

\clearpage


\section{Statement on the Use of Large Language Models (LLMs)}
During the preparation of this manuscript, we use Large Language Models (LLMs) as a general-purpose assistance tool. The primary role of the LLM is to aid in improving the clarity and readability of the text, as well as to accelerate the implementation of our research ideas. Specific applications include: (1) Language and Grammar Correction: Polishing sentence structure, correcting grammatical erros, and refining word choices to enhance the overall quality of the writing. (2) Paraphrasing and Style Refinement: Rephrasing sentences and paragraphs to ensure consistency in tone and style throughout the paper. (3) Code Implementation Assistance: Generating code snippets and providing debugging support to help implement the proposed algorithms and experimental setups.

It should be noted that all core research concepts, experimental design, data analysis, and conclusions are developed exclusively by the human authors. Any content or suggestions generated by the LLM, including code, are critically checked, and substantially edited by the authors to ensure accuracy. The authors take full responsibility for the final content of this paper. 

\section{Baselines Details}
\label{app:baseline}

We mainly compare our method with the following two types of baselines:

\paragraph{(1) LLMs with Search Tools.}
\begin{itemize}[leftmargin=2em]
    \item \textbf{OpenAI GPT-5 w/Search}: The latest OpenAI's GPT model with web search API for research question.
    \item \textbf{Claude-4.1-Sonnet w/Search} The latest Anthropic's reasoning LLM with web search API for research question.
    \item \textbf{DeepSeek-R1 w/ Search}: The DeepSeek's LLM integrated with web search API for research question.
\end{itemize}

\paragraph{(2) Deep Research Agents.}
\begin{itemize}[leftmargin=1.5em]
    \item \textbf{Grok Deep Search}: The xAI's Deep Search applications, powered by the latest Grok model.
    \item \textbf{Perplexity Deep Research}: A commercial AI research assistant integrating multi-step search and analysis, optimized for rapid information aggregation.
    \item \textbf{OpenAI Deep Research}: A multi-step web research agent built on ChatGPT that searches, analyzes, and synthesizes information from multiple sources to produce research-grade reports with citations.
    \item \textbf{Gemini-2.5-Pro Deep Research}: Google’s advanced research agent featuring multi-turn planning, deep web navigation, and multi-source evidence integration.
\end{itemize}

We evaluate these baselines directly on their official web applications. For consistency across different systems, we use the following unified prompt template to get the report: 

\begin{tcolorbox}[
    colframe = red!50!black,            
    colback = gray!5!white,             
    coltitle = white,                   
    coltext = black,                    
    fonttitle = \bfseries,              
    title = PROMPT,         
    boxrule = 1pt,                      
    arc = 1mm,                          
    width = \linewidth,                 
    left = 5pt,                         
    right = 5pt,                        
    top = 5pt,                          
    bottom = 5pt,                        
]
Please help me write a detailed research report on the corporate finance of \{topic\}, which should be rich in both text and charts. Give me the standardized citations at the end of the report (including serial numbers and corresponding references).
\end{tcolorbox}

\section{Implementation Details of \ours{}}
\label{app:implementation}

\paragraph{Backbone} For Multi-source Data Collection Agent, Deep Search Agent and Data Analysis Agent, we use the \texttt{DeepSeek-V3} as the backbone model. For Report Generation Agent, we use \texttt{DeepSeek-R1} as the backbone model. The maximum input length is 81,920 tokens, and the maximum output length is 16,384 tokens. 

\paragraph{Data Collection} We implement the financial api tool based on akshare~\footnote{https://github.com/akfamily/akshare} package in Python. For web search, we use the Google Search API, with the region set to China and the number of retrieved results fixed at the top 10. For web content acquisition, we employ Playwright~\footnote{https://playwright.dev/} to simulate a browser for webpage content extraction.

\paragraph{Retrieval} We use \texttt{Qwen3-Embedding-0.6B} to generate embeddings for data and CoA segments. Then we use the cosine similarity to select the relevant data and CoA segments for each section. 

\paragraph{Iterative Vision-Enhanced Mechanism} We use the \texttt{Qwen2.5-VL-72B} as the critic vision-language model in the chart generation stage. To balance effectiveness and cost, we perform three iterations of the critic process.

\paragraph{Ablation Study}
We conduct ablation study on $5$ company questions, which includes: Cambricon Technologies, Li Auto-W, Pop Mart, 3SBio, China Mobile. Some variants are as follows:

\begin{itemize}[leftmargin=1.5em]
    \item \textbf{w/o Iteration Vision-Enhanced Mechanism} We remove the iterative refinement process and plot charts in a single pass.
    \item \textbf{w/o Two-Stage Writing Framework} We only concatenate the CoA segments to output the final report.
    \item \textbf{w/o Dynamic Search-Enhanced Strategy} We remove the Dynamic Search-Enhanced Strategy from the Data Collection and Report Generation process.
\end{itemize}

\section{Construction of the Financial Report Generation Benchmark}
\paragraph{Questions} 
We select the most popular five A-share companies, five Hong Kong-stock companies, and ten representative industries from \url{https://www.djyanbao.com} as the benchmark research questions. These companies and industries cover a diverse set of market sectors and provide a comprehensive foundation for evaluating the effectiveness of deep research systems.

\paragraph{Golden Referenced Report} 
To establish human expert-level benchmark, we collect the latest equity and industry research reports from well-known Chinese securities firms, as shown in Table~\ref{tab:golden_reports}. These golden references cover both company-level and industry-level analyses across A-shares, Hong Kong stocks, and major industries.

\begin{table}[ht]
\centering
\caption{Golden Referenced Reports from Chinese Securities Firms}
\label{tab:golden_reports}
\begin{tabularx}{\textwidth}{lXl}
\toprule
\textbf{Market} & \textbf{Company / Industry} & \textbf{Securities Firm} \\
\midrule
A-shares 
& SMIC (688981) & Soochow Securities \\
& Cambricon Technologies (688256) & Donghai Securities \\
& China Mobile (600941) & Zhongtai Securities \\
& Skshu Paint (603737) & Huatai Securities \\
& Yiwu China Commodities City (600415) & Guolian Minsheng Securities \\
\midrule
Hong Kong Stocks 
& Pop Mart (09992) & Zhongtai Securities \\
& SenseTime (00020) & Zhongtai Securities \\
& Li Auto-W (02015) & Huayuan Securities \\
& 3SBio (01530) & Huatai Securities \\
& UBTECH Robotics (09880) & Guohai Securities \\
\midrule
Industries 
& Semiconductor Industry & Kaiyuan Securities \\
& Food \& Beverage Industry & Huachuang Securities \\
& Basic Chemical Industry & Zhongtai Securities \\
& Steel Industry & Orient Securities \\
& Construction \& Decoration Industry & Guosheng Securities \\
& Environmental Protection \& Public Utilities (Controlled Nuclear Fusion) & Huachuang Securities \\
& Light Manufacturing (Durable Consumer Goods) & Guotai Haitong Securities \\
& K12 Education Industry & Guosheng Securities \\
& Media Industry (Short Drama Overseas Expansion) & Soochow Securities \\
& Transportation (Cross-border E-commerce Logistics) & Maigao Securities \\
\bottomrule
\end{tabularx}
\end{table}

\section{Evaluation and Metrics}
\label{app:eval}
We further illustrate the metrics we used for evaluation:

\paragraph{(1) Factual Metrics} Measure the textual quality and factual accuracy of the final report.

\begin{itemize}[leftmargin=1.5em]
    \item \textbf{Core Conclusion Consistency}: Whether the core conclusions in the generated report are consistent with those in the reference report.
    \item \textbf{Textual Faithfulness}: Whether the arguments in the report are properly supported by citations from the reference.
    \item \textbf{Text-Image Coherence}: Whether the report integrates images into the discussion, and whether the textual and visual descriptions align.
\end{itemize}

\paragraph{(2) Analysis Effectiveness} Measure whether the financial report provides sufficient information and insights for investors.
\begin{itemize}[leftmargin=1.5em]
    \item \textbf{Information Richness}: The number of distinct information points included in the report.
    \item \textbf{Coverage}: The extent to which key information from the golden reference report is covered.
    \item \textbf{Analytical Insight}: Whether the report provides critical analysis, original insights, and forward-looking recommendations.
\end{itemize}

\paragraph{(3) Presentation Quality} Measure the presentation quality of the final report.
\begin{itemize}[leftmargin=1.5em]
    \item \textbf{Structural Logic}: The logical organization of each section and the overall structural soundness of the report.
    \item \textbf{Language Professionalism}: Whether the language conforms to financial terminology, using the golden report as a reference.
    \item \textbf{Chart Expressiveness}: The effectiveness of charts in supporting the narrative, including their informativeness and aesthetic quality.
\end{itemize}

\paragraph{Evaluation Process} We adopt \texttt{Gemini-2.5-Pro} as the backbone evaluation model. To ensure fair comparison across reports, we employ a list-wise evaluation strategy, where the model is provided with all candidate reports along with the golden reference report and assigns scores accordingly. 
The nine metrics mentioned above can be divided into two parts, one is unrelated to the golden report and the other is related to the golden report. For these two types, we have designed two types of prompts, which are listed below.

\begin{tcolorbox}[
    enhanced jigsaw,
    breakable,
    pad at break*=1mm,
    colframe = gray!115,       
    colback=gray!5!white,             
    coltitle=white,                   
    coltext=black,                    
    fonttitle=\bfseries,              
    title=Evaluation Instruction for Golden Report Irrelevant Metrics,  
    boxrule=1pt,                      
    arc=2mm,                          
    width=\linewidth,                 
    left=7pt,                         
    right=7pt,                        
    top=5pt,                          
    bottom=5pt                        
]
\fontsize{8.5pt}{10pt}\selectfont
\# [TASK]

Your task is to act as an expert financial analyst and editor. You will perform a rigorous, **comparative evaluation** of a list of financial research reports. Your goal is to produce a structured critique for each report based on how effectively it addresses the central **Research Question**, using the provided **Golden Standard Report** as a quality benchmark.

\# [INPUTS]

  * **Research Question:** {{Research Question}}
  * **Golden Standard Report:** Given in file format, the one starting with 'golden' is the 'golden standard report'
  * **Reports to Evaluate:** Reports 

\# [EVALUATION METHODOLOGY]

To ensure fairness and accuracy, you must follow this three-step process for **each report** in the `Reports to Evaluate` list:

1.  **Step 1: Establish the Benchmark (Internal Thought Process)**

      * For each of the six evaluation dimensions, first thoroughly analyze the **Golden Standard Report**. Identify its key characteristics, depth, and quality to create a mental benchmark for what constitutes a high-quality, professional report (which corresponds to a score of 7).

2.  **Step 2: Comparative Analysis (Internal Thought Process)**

      * Now, analyze the report currently being evaluated. For each dimension, find concrete evidence (e.g., specific quotes, data points, chart quality, structural features).
      * **Directly compare** this evidence against the benchmark established in Step 1. Note where the report meets, exceeds, or falls short of the Golden Standard.

3.  **Step 3: Score and Justify (Final Output Generation)**

      * Based on the comparison in Step 2, assign a score from 1 to 10 for the dimension, following the `Benchmark-Based Scoring` rules below.
      * Write a **concise, one-sentence rationale** that justifies your score by referencing your comparative findings.

\# [SCORING GUIDELINES]

Adhere strictly to these principles to maintain objectivity:

  * **Benchmark-Based Scoring:**

      * **The Golden Standard Report is the benchmark for a score of 7.**
      * A report demonstrating a **similar level of quality**, depth, and execution as the Golden Standard on a specific dimension should receive a score of **7**.
      * Scores of **8-10** are reserved for reports that **demonstrably exceed** the Golden Standard in that dimension (e.g., providing deeper insights, more comprehensive data, or superior visualizations).
      * Scores of **1-6** indicate that the report **falls short** of the Golden Standard's quality in that dimension, with the score reflecting the degree of the gap.

  * **Justification for Extremes:** Scores of **9-10** (exceptional) or **1-2** (critically flawed) require a particularly strong and specific justification in the rationale.

\# [EVALUATION FRAMEWORK and CRITERIA]

\#\#\# **Dimension 1: Information Richness (Score 1-10)**

  * **Definition:** Measures the concentration of substantive, verifiable facts and data points relevant to the research question, while minimizing filler content.

\#\#\# **Dimension 2: Textual Faithfulness (Score 1-10)**

  * **Definition:** Measures whether significant claims, data, and forecasts are verifiably supported by provided "References / Data Sources".

\#\#\# **Dimension 3: Text-Image Coherence (Score 1-10)**

  * **Definition:** Assesses if charts and tables are consistent with the text and if the text provides meaningful interpretation that supports the core analysis.

\#\#\# **Dimension 4: Analytical Insight (Score 1-10)**

  * **Definition:** Evaluates the quality of the analysis, focusing on critical thinking, original insights, and actionable, forward-looking conclusions that directly address the research question.

\#\#\# **Dimension 5: Structural Logic (Score 1-10)**

  * **Definition:** Measures the structural integrity and logical flow of the argument, assessing if the report builds a clear and compelling case from evidence to conclusion.

\#\#\# **Dimension 6: Chart \& Table Expressiveness (Score 1-10)**

  * **Definition:** Focuses on the quality of data visualizations themselves—their clarity, ability to reveal patterns, and effectiveness in communicating key information.

\# [OUTPUT FORMAT]

Provide your evaluation in the following strict JSON format. **For each score, you must provide a brief, one-sentence rationale.** Do not add any conversational text outside of this structure. Use the file name of each report as its report id.

Now start your evaluation of the given reports. Carefully read each report and give a score.
\end{tcolorbox}

\begin{tcolorbox}[
    enhanced jigsaw,
    breakable,
    pad at break*=1mm,
    colframe = gray!115,       
    colback=gray!5!white,             
    coltitle=white,                   
    coltext=black,                    
    fonttitle=\bfseries,              
    title=Evaluation Instruction for Golden Report Relevant Metrics,  
    boxrule=1pt,                      
    arc=2mm,                          
    width=\linewidth,                 
    left=7pt,                         
    right=7pt,                        
    top=5pt,                          
    bottom=5pt                        
]
\fontsize{8.5pt}{10pt}\selectfont
**[ROLE]**
You are an expert financial analyst and editor, specializing in the comparative analysis of research reports.

**[TASK]**
Your task is to rigorously evaluate a list of **Generated Reports** by comparing each one against a **Benchmark Report** (a professionally written 'gold standard'). You will assess each Generated Report's quality across three key dimensions on a scale of 1 to 10, producing a structured JSON output with scores and justifications.

**[INPUTS]**

1.  **`Benchmark Report`**: A high-quality, professional research report that serves as the "gold standard" for this evaluation. All comparisons should be made against this document. The file name of the benchmark report begins with "golden\_".
2.  **`Generated Reports`**: A list of one or more reports to be evaluated against the Benchmark Report.
3.  **`Report ID`**: An identifier for each Generated Report. Use the file name as the report ID.

**[EVALUATION METHODOLOGY]**

To ensure fairness and accuracy, you must follow this three-step process for **each Generated Report**:

1.  **Step 1: Establish the Benchmark (Internal Thought Process)**

      * For each of the three evaluation dimensions, first thoroughly analyze the **Benchmark Report**. Identify its key characteristics, depth, and quality to create a mental benchmark for what constitutes a score of **7**.

2.  **Step 2: Comparative Analysis (Internal Thought Process)**

      * Now, analyze the Generated Report. For each dimension, find concrete evidence (e.g., specific conclusions, data points included/omitted, linguistic style).
      * **Directly compare** this evidence against the benchmark established in Step 1. Note where the report meets, exceeds, or falls short of the Benchmark Report.

3.  **Step 3: Score and Justify (Final Output Generation)**

      * Based on the comparison in Step 2, assign a score from 1 to 10 for the dimension, following the `SCORING GUIDELINES` below.
      * Write a **concise, one-sentence rationale** that justifies your score by referencing your comparative findings.

**[SCORING GUIDELINES]**

Adhere strictly to these principles to maintain objectivity:

  * **Benchmark-Based Scoring:**
      * **The Benchmark Report is the standard for a score of 7.**
      * A report demonstrating a **similar level of quality**, depth, and execution as the Benchmark Report on a specific dimension should receive a score of **7**.
      * Scores of **8-10** are reserved for reports that **demonstrably exceed** the Benchmark Report in that dimension (e.g., providing a more nuanced conclusion, broader data coverage, or more sophisticated language).
      * Scores of **1-6** indicate that the report **falls short** of the Benchmark Report's quality in that dimension, with the score reflecting the degree of the gap.
  * **Justification for Extremes:** Scores of **9-10** (exceptional) or **1-2** (critically flawed) require a particularly strong and specific justification in the rationale.

**[EVALUATION FRAMEWORK \& CRITERIA]**

\#\#\# **Dimension 1: Core Conclusion \& Data Consistency (Score 1-10)**

  * **Definition:** Measures the alignment of the Generated Report's core thesis, key arguments, and supporting data points with those presented in the Benchmark Report.

\#\#\# **Dimension 2: Information Coverage (Score 1-10)**

  * **Definition:** Assesses the extent to which the Generated Report includes the key information points, topics, and analytical angles present in the Benchmark Report.

\#\#\# **Dimension 3: Professional Language \& Tone (Score 1-10)**

  * **Definition:** Evaluates the linguistic quality of the Generated Report, using the Benchmark Report's writing style, tone, and vocabulary as the standard for professional financial analysis.

**[OUTPUT FORMAT]**
Provide your evaluation in the following strict JSON format. For each score, you must provide a brief, one-sentence rationale that explains the score relative to the benchmark. Do not add any conversational text outside of this structure.

Now start your evaluation of the given reports. Carefully read each report and give a score.
\end{tcolorbox}

\clearpage
\section{A Case of Company Research Question}
To demonstrate the practical application of our system, this section shows the case of \textbf{SenseTime Technology (0020.HK)}, a leading artificial intelligence company in China. 

We present the collecting tasks of the Data Collection process in Table~\ref{tab:data_collection}, and an analytical tasks of the Data Analysis process in Table~\ref{tab:data_analysis}.

\begin{table*}[htbp]
    \centering
    \caption{The predefined and brainstormed data collection tasks.}
    \fontsize{9pt}{11pt}\selectfont
    \begin{tabular}{p{0.98\linewidth}}
    \toprule
\rowcolor{gray!20}\textbf{Data Collection} \\
\midrule
\begin{lstlisting}[basicstyle=\ttfamily\footnotesize, breaklines=true]
Predefined Tasks:
"company": [
    {"name": "Balance Sheet"},
    {"name": "Income Statement"},
    {"name": "Cash Flow Statement"},
    {"name": "Basic Stock Information"},
    {"name": "Shareholder Structure"},
    {"name": "Stock Price"},
    {"name": "Stock-related Financial Data"},
    {"name": "CSI 300 Daily Index Data"},
    {"name": "Hang Seng Daily Index Data"},
    {"name": "NASDAQ Daily Index Data"},
    {"name": "Investment Rating", 
    "description": "Collect analyst investment ratings and target prices from major securities firms (e.g., CITIC Securities, CICC). Select the most frequent rating across institutions."
    },
    {"name": "Price-to-Sales Ratio (PS)",  
    "description": "Collect the company's PS data (PS = Stock Price / Sales per Share)."
    },
    {"name": "Return on Equity (ROE, %)", 
    "description": "Collect the company's ROE data, defined as (Net Income Attributable to Shareholders / Equity) or equivalently (Net Profit Margin x Asset Turnover x Equity Multiplier)."
    },
    { "name": "Competitor Data", 
    "description": "Collect comparable data from 5-8 major peer companies, including financial and operational indicators."
    },
]

Brainstormed Tasks: [
    {"name": "Growth Data for AI Subfields in 2024"},
    {"name": "The Development History of The SenseTime and Main Business"}
]

\end{lstlisting} \\

    \bottomrule
    \end{tabular}
    \label{tab:data_collection}
\end{table*}

\begin{table*}[htbp]
    \centering
    \caption{The illustration of data analysis process and CoA output of a task.}
    \fontsize{9pt}{11pt}\selectfont
    \begin{tabular}{p{0.98\linewidth}}
    \toprule
\rowcolor{gray!20}\textbf{Data Analysis (We show a predefine task of data analysis, and its CoA output.)} \\
\midrule
\textbf{Predefined Task 1:} Company Development History and Main Business Analysis \\
\textbf{Required Data:} Years and corresponding company achievements. \\
\textbf{Description:} Organize the company’s key milestones by year (e.g., founding, expansions, product launches, strategic shifts, and M\&A) and present them as a timeline with annotated events. Also summarize its current core business scope, including main products, business segments, and market positioning. \\
\midrule
\textbf{CoA Output:} \\
\\
\textbf{\#\# SenseTime: Evolution from Computer Vision Pioneer to AI Platform Enterprise} \\

\textbf{\#\#\# Executive Summary} \\
As a global leading artificial intelligence software company, SenseTime has grown into an AI giant valued at over USD 10 billion since its founding in 2014. With the mission of ”persisting in originality and enabling AI to advance human progress", the company has built a comprehensive AI ecosystem covering four major business segments: Smart Business, Smart City, Smart Life, and Smart Auto, through 11 years of continuous innovation and business expansion \textcolor{orange}{[Source: SenseTime Group Company Profile]}. By 2024, SenseTime employed 3,756 people, with operations spanning over 20 cities worldwide and R\&D centers in Hong Kong, Shanghai, Beijing, and Shenzhen \textcolor{orange}{[Source: SenseTime Basic Information]}. \\

The company’s growth trajectory reflects a typical path of technology firms: from technical breakthroughs, to productization, then to platformization and ecosystemization. Its IPO on the Hong Kong Stock Exchange in 2021 (Ticker: 0020.HK) marked the beginning of a new capital-intensive phase, while the launch of the "Riyadh New" large model system in 2023 signified its strategic pivot toward generative AI \textcolor{orange}{[Source: Development History Search Data]}. Currently, SenseTime reports annual revenue of RMB 3.772 billion, maintaining technological leadership in computer vision and artificial intelligence \textcolor{orange}{[Source: Stock 0020 Income Statement Data]}. \\
\\
\textbf{\#\#\# Development History Analysis} \\

\textbf{Early Stage of Founding and Technology Accumulation (2014–2016)} \\

Founded on October 15, 2014, in the Cayman Islands by the CUHK Multimedia Lab team, SenseTime initially focused on core R\&D in computer vision and deep learning \textcolor{orange}{[Source: SenseTime Basic Information]}. In 2015, it launched its first facial recognition algorithm and completed its first funding round, laying the foundation for subsequent commercialization. By 2016, it had become the world’s most valuable AI unicorn, reflecting strong capital market recognition. \\
\\
\textcolor{darkblue}{\textbf{@import "SenseTime Development Timeline (2014-2025), horizontal axis = year, vertical axis = milestone importance, annotated with key events (milestone\_df)"}} \\
\\
\textbf{Productization and Market Expansion (2017–2019)} \\

In 2017, the launch of the SenseAR augmented reality platform marked the transition from pure research to productization. By 2018, it had completed Series C financing with a valuation exceeding USD 4.5 billion. In 2019, it rolled out Smart City solutions, partnering with key domestic cities and expanding its scope from technology output to integrated solutions. \\

\textbf{Platformization and Capitalization (2020–2022)} \\

In 2020, the release of the SenseCore AI Infrastructure introduced the concept of industrializing AI, aiming to reduce the cost of AI production factors. On December 30, 2021, SenseTime was listed on the Hong Kong Stock Exchange Main Board, raising about HKD 6 billion \textcolor{orange}{[Source: SenseTime IPO Search Data]}. In 2022, it launched the ”Riyadh New" foundation model system, officially entering the generative AI space and advancing toward multimodal development. \\

\textbf{Ecosystemization and Deepening Applications (2023–2025)} \\

In 2023, the company released the ”SenseChat" large language model, further completing its model matrix. In 2024, its Smart Auto business grew rapidly, with partnerships across major automakers. By 2025, SenseTime continued to optimize its large model technologies and achieved deep application deployments in healthcare, finance, and manufacturing. \\
\\
\textbf{\#\#\# Core Business Structure Analysis} \\

    \bottomrule
    \end{tabular}
    \label{tab:data_analysis}
\end{table*}

\begin{table*}[!t]
    \centering
    
    \fontsize{9pt}{11pt}\selectfont
    \begin{tabular}{p{0.98\linewidth}}
    \toprule
\textbf{Smart Business (35\%)} \\

Core Products: SenseFoundry Enterprise Ark, AI Content Generation Platform
Market Positioning: Provides AI solutions for enterprise digital transformation, focusing on finance, retail, and manufacturing. This is the main revenue driver, contributing about 35\% of revenue in 2024 \textcolor{orange}{[Source: Business Segment Data]}. \\

\textbf{Smart City (30\%)} \\

Core Products: City Ark, traffic management, and surveillance systems \\
Market Positioning: Provides smart governance and public service solutions. Established partnerships with major cities, gaining competitive advantages in smart transportation and public safety. \\

\textbf{Smart Life (20\%)} \\

Core Products: SenseMARS Metaverse Platform, mobile AI applications \\
Market Positioning: Consumer-facing AI applications and metaverse ecosystem. Focuses on AR/VR and mobile internet scenarios, with strong growth potential. \\

\textbf{Smart Auto (15\%)} \\

Core Products: SenseAuto Platform, autonomous driving solutions \\
Market Positioning: Provides intelligent driving and V2X systems for automakers. This segment grew rapidly in 2024, becoming a new growth engine \textcolor{orange}{[Source: Smart Auto Data]}. \\
\\
\textcolor{darkblue}{\textbf{@import "SenseTime 2024 Revenue Breakdown Pie Chart, showing four business segments (business\_df)"}} \\
\\
\#\#\# \textbf{Technical Strength and Competitive Advantages} \\

SenseTime possesses deep academic roots and originality in AI technologies. The company has built full-stack AI capabilities across perception intelligence, decision intelligence, content generation, and content enhancement \textcolor{orange}{[Source: Company Profile]}. \\

By 2024, it had over 8,000 AI-related patents, maintaining leadership in computer vision, large model training, and AI chips \textcolor{orange}{[Source: Patent Data]}. \\

Core Competitive Advantages: \\

	1.	Technological Originality: Commitment to independent fundamental research and development \\
	2.	Platformization Capability: SenseCore enables AI industrial-scale production \\
	3.	Ecosystem Completeness: Full product portfolio covering four major domains \\
	4.	Talent Density: A world-class AI R\&D team \\
\\
\textbf{\#\#\# Market Position and Prospects} \\

Industry data show SenseTime as a global leader in computer vision, ranking among top players in market share. In 2024, it was positioned in the Leaders Quadrant of the AI maturity curve, highly rated in both innovation and execution \textcolor{orange}{[Source: AI Market Growth Data 2024]}. \\

Future Directions: \\
\ \ \ \ •	Large Model Enhancement: Continuous optimization of the ”Riyadh New" large model system \\
\ \ \ \ •	Vertical Integration: Deepened applications in finance, healthcare, and manufacturing \\
\ \ \ \ •	International Expansion: Strengthening presence in Southeast Asia and the Middle East \\
\ \ \ \ •	Tech Convergence: Driving integration of AI with IoT and blockchain \\
\\
\textcolor{darkblue}{\textbf{@import "SenseTime Revenue Growth Trend (2020-2024), showing revenue growth from RMB 3.446B to RMB 3.772B (income\_statement)"}} \\
\\
\textbf{\#\#\# Conclusion} \\

Through 11 years of development, SenseTime has evolved from a single computer vision technology provider into a comprehensive AI platform enterprise. Its trajectory reflects a clear progression: technology-driven, product iteration, platform building, and ecosystem expansion. With balanced business segments and strong technological capabilities, SenseTime holds a favorable competitive position in the AI industrialization wave. \\
\\
Looking forward, with continuous AI evolution and expanding applications, SenseTime is poised to play a more significant role in the global AI market. To remain competitive, the company must sustain R\&D investment, deepen vertical applications, and enhance international collaborations in response to intensifying market and technological challenges. \\
    \bottomrule
    \end{tabular}
    \label{tab:generate_data_prompt}
\end{table*}
\newpage
\section{Report Gallery}
We have presented an overview of the report generated by \ ours {} here, and the complete report can be obtained from \url{https://anonymous.4open.science/r/FinSight-6739/}.

\begin{figure*}[htbp]
    \centering
    \includegraphics[width=.9\linewidth]{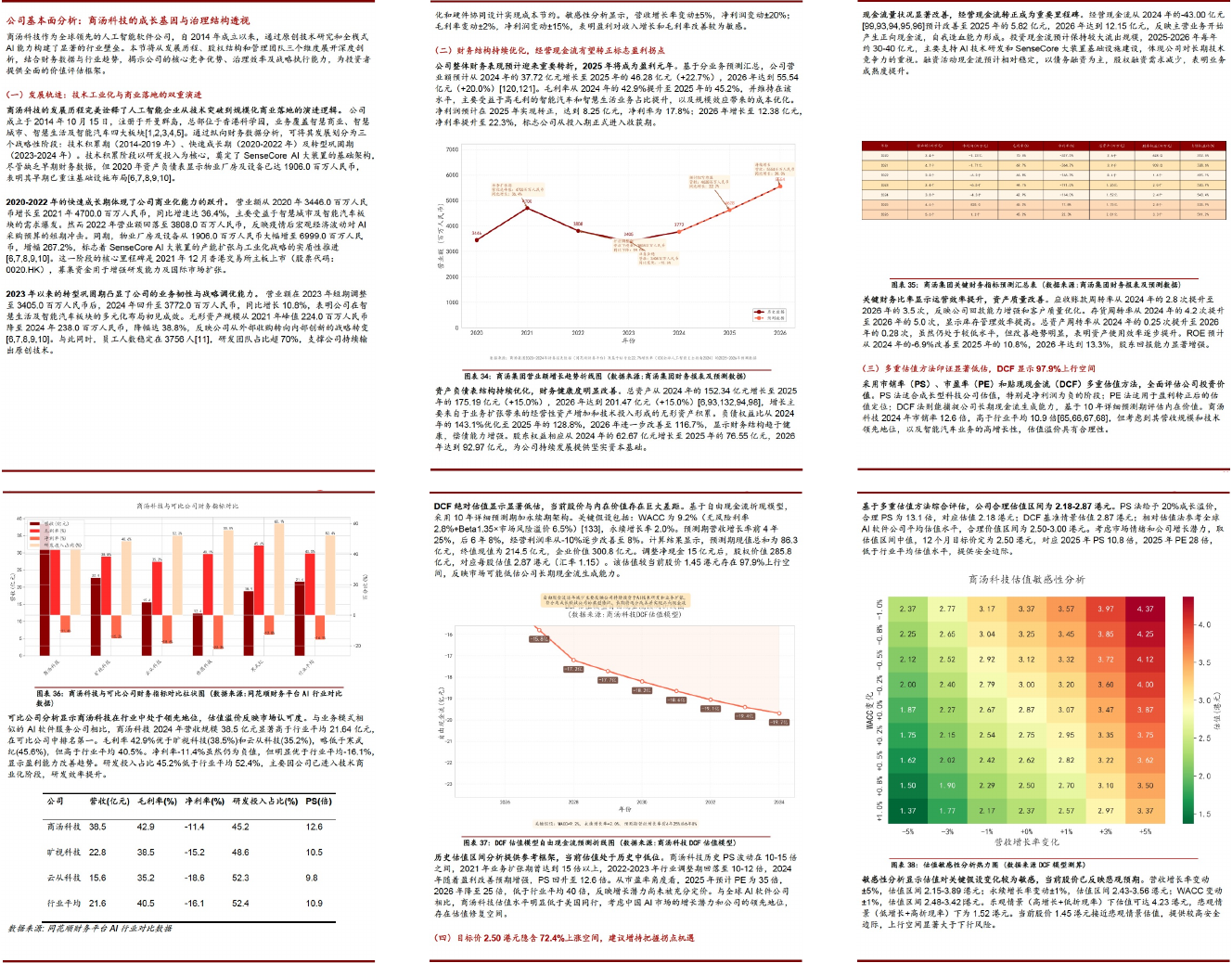}
    \caption{
    The final report of The SenseTime (part).
    }
\end{figure*}

\begin{figure*}[htbp]
    \centering
    \includegraphics[width=.9\linewidth]{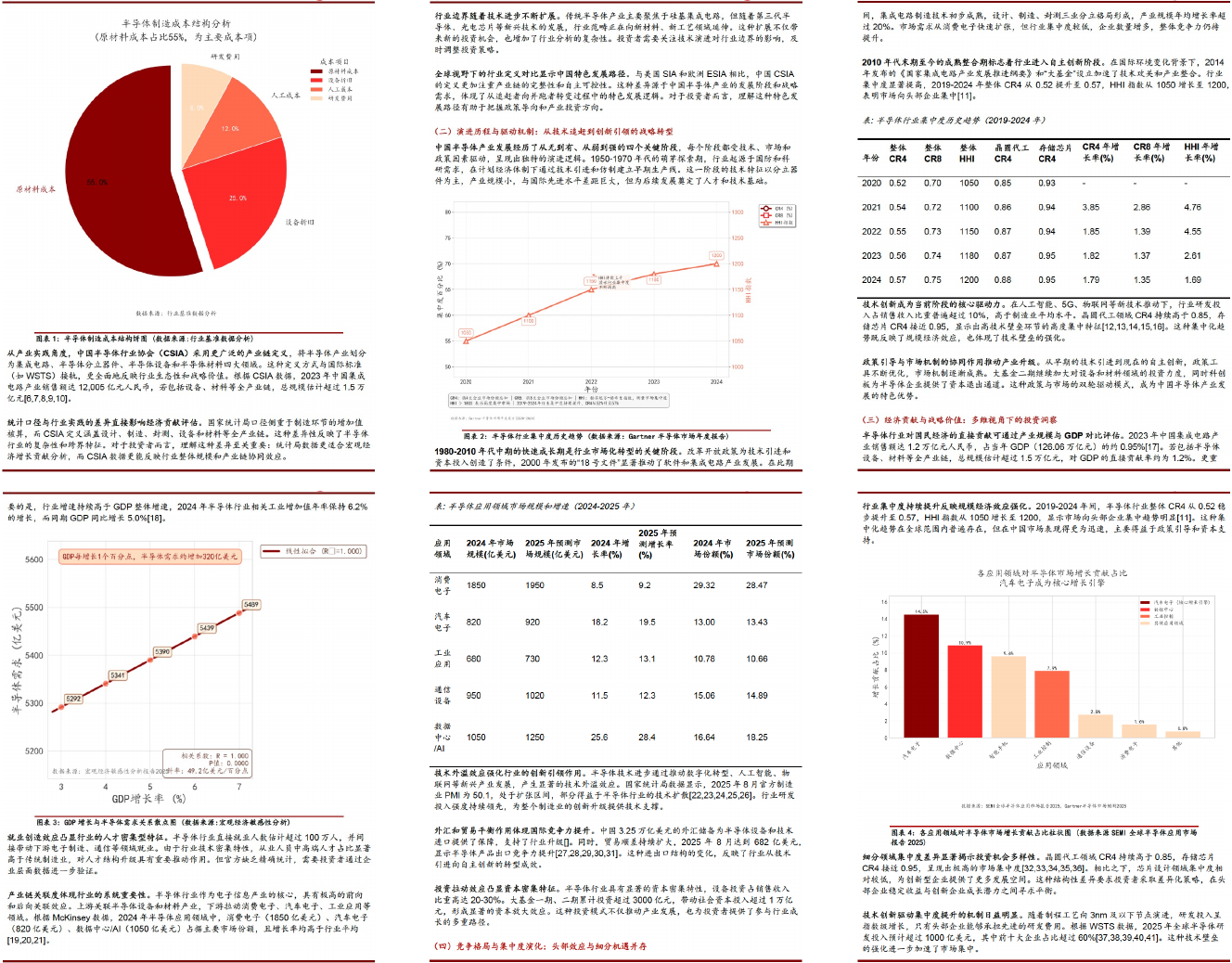}
    \caption{
    The final report of semiconductor industry (part).
    }
\end{figure*}

\begin{figure*}[htbp]
    \centering
    \includegraphics[width=.9\linewidth]{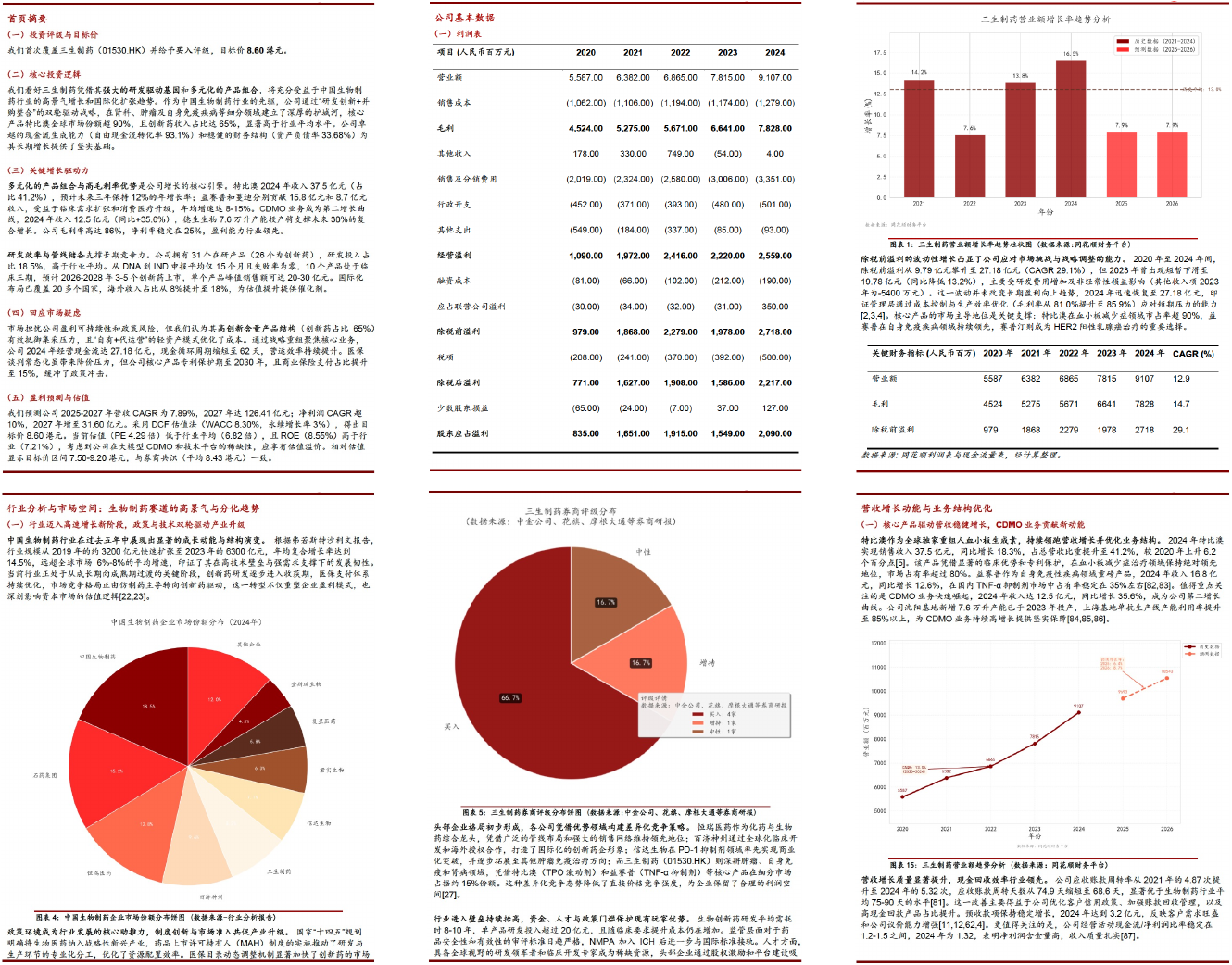}
    \caption{
    The final report of The 3SBio Inc. (part).
    }
\end{figure*}

\begin{figure*}[htbp]
    \centering
    \includegraphics[width=.9\linewidth]{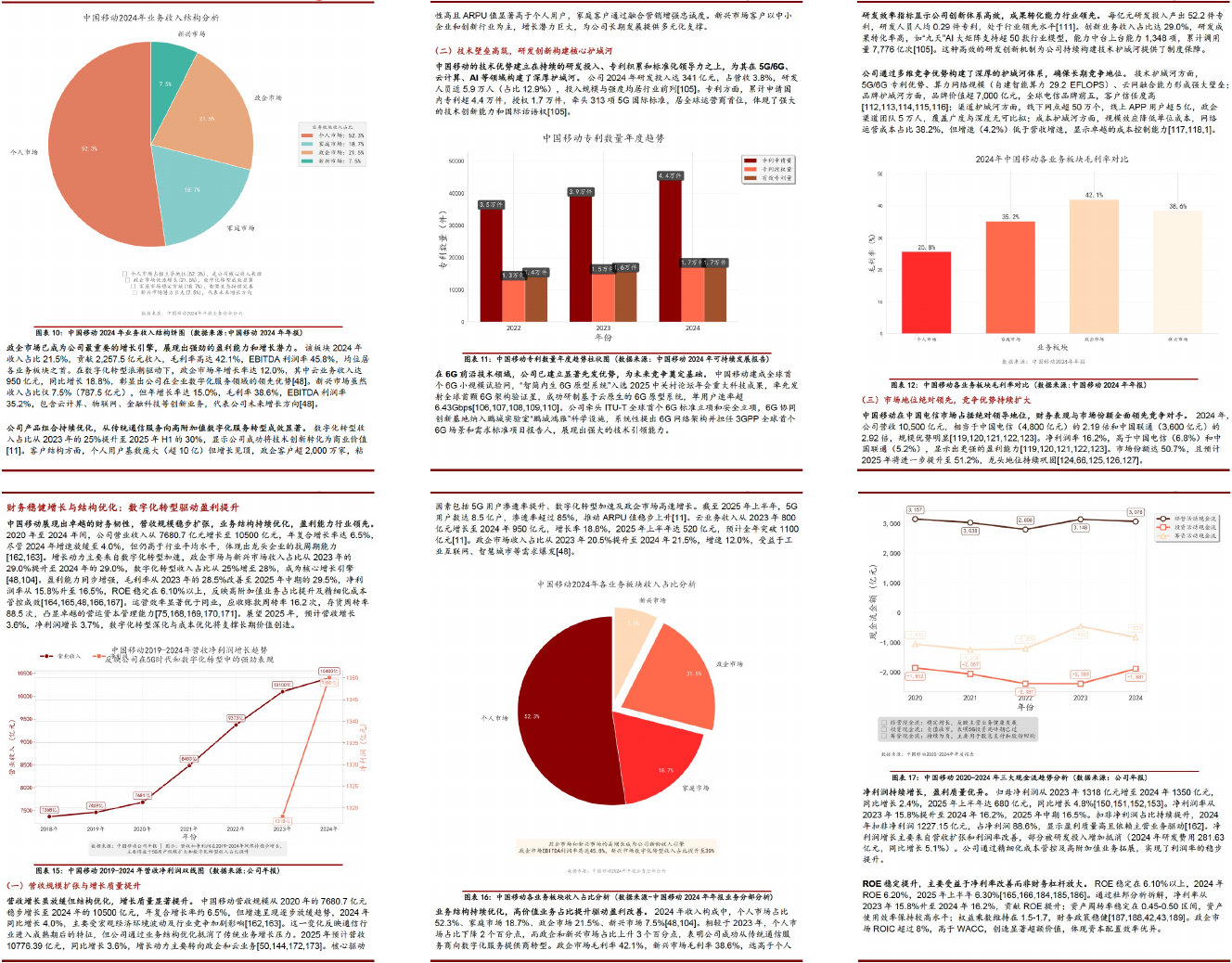}
    \caption{
    The final report of The China Mobile (part).
    }
\end{figure*}

\end{document}